\documentclass{article}

\usepackage[preprint]{neurips_2019}

\usepackage[utf8]{inputenc} 
\usepackage[T1]{fontenc}    
\usepackage{hyperref}       
\usepackage{url}            
\usepackage{booktabs}       
\usepackage{amsfonts}       
\usepackage{nicefrac}       
\usepackage{microtype}      

\usepackage{algorithm}
\usepackage{algorithmicx}
\usepackage{algpseudocode}
\usepackage{amsthm}
\usepackage{amsmath}
\usepackage{mathtools}
\usepackage{enumitem}
\usepackage{xspace}

\usepackage{graphicx}
\usepackage{wrapfig}

\usepackage{pgfplots}
\pgfplotsset{compat=newest}
\usepgfplotslibrary{groupplots}
\usepgfplotslibrary{dateplot}

\newtheorem{theorem}{Theorem}

\usepackage{xr-hyper}

\makeatletter
\newcommand*{\addFileDependency}[1]{
  \typeout{(#1)}
  \@addtofilelist{#1}
  \IfFileExists{#1}{}{\typeout{No file #1.}}
}
\makeatother

\newcommand*{\myexternaldocument}[1]{%
    \externaldocument{#1}%
    \addFileDependency{#1.tex}%
    \addFileDependency{#1.aux}%
}

\myexternaldocument{./supplementary}

\title{All-Action Policy Gradient Methods: \\ A Numerical Integration Approach}

\author{%
  Benjamin Petit\\
  Stanford University \\
  Stanford, CA \\
  \texttt{benpetit@stanford.edu} \\
  \And
  Loren Amdahl-Culleton \\
  Stanford University \\
  Stanford, CA \\
  \texttt{lkac@stanford.edu} \\
  \And
  Yao Liu \\
  Stanford University \\
  Stanford, CA \\
  \texttt{yaoliu@stanford.edu} \\
  \And
  Jimmy Smith \\
  Stanford University \\
  Stanford, CA \\
  \texttt{jsmith14@stanford.edu} \\
  \And
  Pierre-Luc Bacon \\
  Stanford University \\
  Stanford, CA \\
  \texttt{plbacon@cs.stanford.edu}
}

\begin{document}

\maketitle

\begin{abstract}
While often stated as an instance of the likelihood ratio trick \citep{Rubinstein1989}, the original policy gradient theorem \citep{Sutton1999} involves an integral over the action space. When this integral can be computed, the resulting ``all-action" estimator \citep{sutton2001} provides a conditioning effect \citep{Bratley1987} reducing the variance significantly compared to the REINFORCE estimator \citep{Williams1992}. In this paper, we adopt a numerical integration perspective to broaden the applicability of the all-action estimator to general spaces and to any function class for the policy or critic components, beyond the Gaussian case considered by \citet{ciosek2018}. In addition, we provide a new theoretical result on the effect of using a biased critic which offers more guidance than the previous ``compatible features" condition of \citet{Sutton1999}. We demonstrate the benefit of our approach in continuous control tasks with nonlinear function approximation. Our results show improved performance and sample efficiency.
\end{abstract}

\section{Introduction}
\label{intro}
Likelihood ratio (LR) gradient estimators \citep{Aleksandrov1968,Glynn1987,Reiman1989,Rubinstein1989} have been widely used in reinforcement learning \citep{SuttonBarto2018} since the seminal work of \cite{Williams1992} in the class of policy gradient methods \citep{Kimura1995,Kimura1997,Kimura1998,Marbach1998,Sutton1999,Konda2000,Baxter2001}. The popularity of LR methods stems from its ease of implementation \citep{Schulman2015} and its applicability to both discrete and continuous actions spaces \citep{Konda2000}, in the batch or online settings \citep{Baxter2001}.

However, likelihood ratio methods may also suffer \citep{Mania2018} from high variance in the long horizon setting \citep{Lecuyer1991} or when the change of measure fails to have full support \citep{Lecuyer1990}.
Hence, variance reduction techniques \citep{Bratley1987,Lecuyer1994} must usually be used in conjunction with the LR method \citep{Lecuyer1991}.
Three main strategies are employed in practice, namely: 1) the \textit{baseline} approach \citep{Williams1992} 2) by leveraging the Markovian structure \citep{Williams1992,Glynn1995} 3) by chunking a long trajectory into smaller \textit{replications} via a \textit{regenerative} state \citep{Lecuyer1991,Baxter2001,Konda2000} or by truncating the horizon with a smaller discount factor \citep{Baxter2001}.

An intuitive but lesser known variance reduction technique is that of \textit{conditioning}, stemming from \textit{conditional Monte Carlo} methods \citep{Hammersley1964,Bratley1987}. At a high level: if there is a $Y$ such that $\mathbb{E}\left[X\right]=\mathbb{E}\left[\mathbb{E}\left[X|Y\right]\right]$ then by the law of iterated expectation the variance may be reduced by computing $\mathbb{E}\left[X|Y\right]$ separately.
Conditioning is exactly the principle at play behind the so-called \textit{expected} methods \citep{vanSeijen2009,ciosek2018,SuttonBarto2018} in RL (although this connection had never been stated explicitly before).

Interestingly, the very statement of the \textit{policy gradient theorem} by \cite[theorem 1]{Sutton1999} hints clearly at an application of the conditioning technique, yet this approach has not been widely used in practice, nor analyzed properly. In an unfinished paper, \cite{sutton2001}, posit the superiority of this approach in what they call the \textit{all-action} policy gradient estimator and lay out an agenda to show this formally and in practice (but never provided those results). For more than a decade, the \textit{all-action} form has been mostly forgotten in favor of the \textit{single-action} LR variant and was only re-discovered recently in \citep{Allen2017,ciosek2018,Fellows2018}.

In this paper, we provide a first explanation as to why all-action methods may improve the variance of policy gradient estimators by establishing a connection to conditional Monte Carlo methods. Using ideas from the numerical integration literature \citep{Forsythe1977}, we then propose general policy gradient methods capable of implementing all-action estimators in continuous action spaces. More precisely, we investigate the use of quadrature formula and Monte Carlo integration methods for approximating the conditional expectation over actions involved in the policy gradient theorem. This approach is flexible and does not require a specific parameterization of the \textit{actor} or \textit{critic} components as in \cite{ciosek2018}. We show that our perspective applies readily to nonlinear function approximators and can scale to the challenging environments in MuJoCo \citep{Todorov2012}.

We also provide a number of new theoretical results pertaining to the use of a biased critic in policy gradient methods. These new insights contribute to a better understanding of policy gradient methods with general function approximation, beyond the limited scope of the \textit{compatible features} theorem \cite[theorem 2]{Sutton1999} for linear critics. In particular, theorem \ref{thm:biased_SGD} is a general result on the expected dynamics of stochastic gradient ascent for a biased critic. It shows that if the bias term can be controlled, then a good solution may still be obtained. This result mirrors a similar condition for SGD requiring the noise to vanish in the limit \citep{bertsekas2016nonlinear}. In the case of Monte-Carlo integration, theorem \ref{thm:MC_MSE} provides a bound on the mean squared error of the all-action estimator with a learned critic and is shown to decrease as a function of the number of sampled actions. This error is finally compared with that of the LR estimator based on rollouts in theorem \ref{thm:REINFORCE_MSE}.

\section{Background and Notation}
\label{background}
The following presentation is based on the the Markov Decision Process (MDP) framework in the infinite horizon discounted setting \citep{Puterman1994}. We assume a continuous discounted MDP $(\mathcal{S}, \mathcal{A}, P, R, \gamma)$ (more details can be found in the supplementary material).

Policy gradient methods seek to identify an optimal policy by searching within a designated parameterized set of policies by gradient ascent. The concept of optimality in this case is defined with respect to the expected discounted return from a given initial state distribution. We write $\pi^* = \arg\max_{\pi} J(\pi)$ to denote an optimal policy where $J(\pi) \equiv \mathbb{E}_{\tau}\left[G(\tau)\right] \equiv \mathbb{E}_{\tau}\left[\sum_{t=0}^{T-1} \gamma^t r(s_t,a_t) \right]$.
 The policy gradient theorem \citep{Sutton1999} provides an expression for the gradient of expected return with respect to the parameters of a policy. Let $d_{\pi_\theta} \in \mathcal{M}(\mathcal{S})$ \citep{Konda2000} be the \textit{discounted stationary distribution} of state under policy $\pi$, and $J(\theta) \equiv J(\pi)$ when $\pi$ is parameterized by $\theta$. The policy gradient theorem \citep[theorem 1]{Sutton1999} (henceforth referred to as PGT) states that:
\begin{equation}
    \label{eq:originalPG1}
    \nabla_{\theta}J(\theta) = \mathbb{E}_{s \sim d_{\pi_{\theta}}(\cdot)} \left [ \int_{a \in A} \frac{\partial \pi_\theta(a|s)}{\partial \theta} Q^{\pi_\theta}(s,a) da \right ] \equiv \mathbb{E}_{s \sim d_{\pi_{\theta}}(\cdot)} \left [ Z(s,\theta) \right]  \enspace .
\end{equation}
Because the term within the expectation involves an integral over the action space, it is often more convenient to proceed to a \textit{change of measure} \citep{Lecuyer1990} by the likelihood ratio approach:
\begin{equation}
    \label{eq:originalPG2}
    \nabla_{\theta}J(\theta) = \mathbb{E}_{s \sim d_{\pi_{\theta}}(\cdot)} \left [ \mathbb{E}_{a \sim \pi_{\theta}(\cdot|s)} \left[ \frac{\partial \log \pi_\theta(a|s)}{\partial \theta} Q^{\pi_\theta}(s,a) \right] \right ]
    \equiv \mathbb{E}_{s} \left [ \mathbb{E}_{a \sim \pi_{\theta}(\cdot|s)} \left[ z(s,a,\theta) \right] \right ] \enspace .
\end{equation}
In this paper, we assume that $\nabla \pi_{\theta} / \pi_{\theta}$ is uniformly bounded to ensure that the change of measure are well defined. Policy search methods \citep{Szepesvari2010} -- such as REINFORCE \citep{Williams1992} -- use sample future returns $G_t = \sum_{s=t}^{T} \gamma^{s-t} r_s$ (rollouts) in an estimator of the form:
\begin{equation}\label{eq:REINFORCE}
    \nabla_{\theta}J(\theta) \simeq \frac{1}{T} \sum_{t=0}^{T-1} \nabla_\theta \log \pi_\theta(a_t|s_t) G_t = \frac{1}{T} \sum_{t=0}^{T-1} \hat{z}(s_t, a_t, \theta) \enspace ,
\end{equation}
whereas actor-critic methods \citep{Sutton1984} maintain an approximation of $Q^{\pi_\theta}$ in a two-timescale manner \citep{Konda2000}.

\section{All-Action Estimators and Conditional Monte Carlo Methods}

By a change of measure via the log trick, we have seen that the inner integral in \eqref{eq:originalPG1} can be transformed into an expectation which can be sampled from along the stationary distribution. While sampling a single action for every state visited along a trajectory would suffice to obtain an unbiased estimator, a better approximation of the inner expectation term is obtained by sampling a larger number of actions. \cite{sutton2001} refer to the former type methods as the \textit{single-action} ones whereas the latter are called \textit{all-action methods}. As shown below, all-action methods are preferable due to their variance reduction effect. However, when the action space is large (or continuous) it may become intractable to compute the inner expectation exactly unless an analytical expression is known apriori as in \citep{ciosek2018}. We address this issue using numerical integration methods.

The variance reduction brought by all-action estimators can be simply understood using the law of total variance \citep{Bratley1987}. Let $\hat{Z}(s)$ be a random variable such that $\mathbb{E}\left[ \hat{Z}(s)|s\right] = Z(s, \theta)$ (an unbiased estimator of the policy gradient)
then the variance of $\hat{Z}(s)$ is:
\begin{equation}\label{eq:total_variance}
    \text{Var} \left[ \hat{Z}(s) \right] = \text{Var}_s \left[ Z(s,\theta) \right] + \mathbb{E}_s \left[ \text{Var} \left[ \hat{Z}(s) | s \right]\right] \enspace .
\end{equation}
The first term corresponds to the variance due to the sampling of actions when computing $Z(s)$ for a given state $s$ while the second one is attributed to the sampling of states. Because the variance is nonnegative for every $s$ in the second term, we have $\text{Var} \left[ \hat{Z}(s) \right] \geq \text{Var}_s \left[ Z(s,\theta) \right]$. This is the core idea behind the so-called \textit{conditional Monte-Carlo} methods \citep{Hammersley1964} and the method of \textit{conditioning} as a variance reduction technique \citep{Bratley1987}.

Because the inner conditional expectation in \eqref{eq:originalPG2} involves the action-value function $Q^{\pi_\theta}$ (unknown in the model-free setting), some algorithmic considerations are needed to implement this idea. When an arbitrarily resettable simulator is available, a potential solution consists in sampling a rollout for every given $(s,a)$ pair. While unbiased, we dismiss this approach due to its high computational cost and lack of generality in domains where offline batch data is available. The solution put forward in this paper consists in using function approximation methods to estimate $Q^{\pi_\theta}$ separately in an actor-critic fashion \citep{Sutton1984} and then numerically integrate the resulting approximate quantity.

\section{Problem Formulation}\label{sec:problem}

In the absence of special structure \citep{ciosek2018}, the main challenge in implementing all-action estimators lies in the intractability of computing the inner expectation in \eqref{eq:originalPG2} for general action spaces. In this paper, we propose two approaches to tackle this problem: numerical quadrature rules and Monte Carlo integration. We aim at deriving efficient estimators of the form:
\begin{equation}
    \nabla J(\theta) \simeq \frac{1}{T} \sum_{t=0}^{T-1} \hat{Z}^{\hat{A}_{\theta}}(s_t) \enspace,
\end{equation}
for a given trajectory $\tau$ and approximate advantage function $\hat{A}_{\theta}$ \citep{baird1993}.

\textbf{Quadrature Formula:} With fixed-grid quadrature methods \citep{Forsythe1977}, $N$ evenly-spaced points are chosen apriori over the range of action space $\{a_1,...,a_N\}$ and we define:
\begin{equation}
    \hat{Z}^{\hat{A}_{\theta}}_{FG}(s,\theta) = \frac{\epsilon}{N} \sum_{i=1}^{N} \frac{\partial \pi_{\theta}(a_i|s)}{\partial \theta} \hat{A}_{\theta}(s,a_i) = \frac{\epsilon}{N} \sum_{i=1}^{N} \hat{z}^{\hat{A}_{\theta}}(s,a_i,\theta) \enspace,
\end{equation}
where $\epsilon$ is the step size and $\hat{z}^{\hat{A}_{\theta}}(s,a,\theta) = \nabla_{\theta} \log \pi_{\theta}(a|s) \hat{A}_{\theta}(s,a)$. These methods generalize to higher dimensions but the number of actions required increases exponentially with the dimension of $\mathcal{A}$. We present a policy update subroutine that uses trapezoidal integration in appendix.

\textbf{Monte Carlo Integration:} While numerical quadrature methods are efficient in low dimensional settings, they display exponential sample complexity as the dimension increases. In higher dimensions, Monte Carlo integration methods offer a performance advantage \citep{evans2000} while being easier to analyze \citep{Glynn1988} and implement. Using the change of measure approach in \eqref{eq:originalPG2}, we derive the all-action Monte Carlo integration estimator from:
\begin{equation}
    Z(s,\theta) = \mathbb{E}_{a \sim \pi_{\theta}(\cdot|s)} \left[ z^{A_{\theta}}(s,a,\theta) \right] \simeq \mathbb{E}_{a \sim \pi_{\theta}(\cdot|s)} \left[ \hat{z}^{\hat{A}_{\theta}}(s,a,\theta) \right] \enspace,
\end{equation}
which we can then approximate from samples using $N_S$ iid samples from $\pi_{\theta}(\cdot|s)$:
\begin{equation}\label{eq:MCPG}
    \hat{Z}^{\hat{A}_{\theta}}_{N_S} (s) = \frac{1}{N_S} \sum_{k=1}^{N_S} \nabla_{\theta} \log \pi_{\theta}(a_k|s) \hat{A}_{\theta}(s,a_k) \enspace .
\end{equation}

\section{Mean Squared Error Analysis}\label{s:analysis}

The all-action estimators presented in section \ref{sec:problem} are cast under the actor-critic framework \citep{Sutton1984} in which an approximation of $Q^{\pi_\theta}$ and the corresponding advantage term are maintained separately. In this section, we investigate how the use of an approximate critic in the all-action estimator affects our ability to estimate the true policy gradient.  More specifically, we provide a bound in theorem \ref{thm:MC_MSE} on the mean squared error of the gradient estimator computed by Monte-Carlo integration, compared to the unbiased LR (REINFORCE) estimator.

\begin{theorem}
\label{thm:MC_MSE}
Assume that for all $s \in \mathcal{S}$ and $a\in\mathcal{A}$, $\log \pi_\theta(s,a)$ is a $\sqrt{M}$-Lipschitz function of $\theta \in \mathbb{R}^d$, i.e.
$
\forall a\in A,\ \forall s\in S,\ ||\nabla_{\theta}\log\pi_{\theta}\left(a|s\right)||_2^2 \leq M \in \mathbb{R} \enspace .
$
Furthermore, let $L^{MC}_{N_S} = \mathbb{E} \left[ || \hat{Z}^{\hat{A}_{\theta}}_{N_S}(s) - \nabla J(\theta)||^2 \right]$ be the MSE (taken w.r.t. both the state and the sampled actions) of the Monte Carlo integration estimate $\hat{Z}^{\hat{A}_{\theta}}_{N_S}(s)$ estimator (Eq. \ref{eq:MCPG}) to the true policy gradient. We have:
\begin{equation} \label{eq:MSE_bound}
    L^{MC}_{N_{S}} \leq M L_{\hat{A}_{\theta}} + \left( L + M  L_{\hat{A}_{\theta}} d\right)/N_S\enspace,
\end{equation}
where $L_{\hat{A}_{\theta}} = \mathbb{E} \left[\left(\hat{A}_{\theta}(s,a)-A_{\theta}(s,a)\right)^{2}\right]$ is the MSE of the advantage estimate and $L = MSE \left[ \nabla_{\theta} \log \pi_{\theta} (a|s) A_{\theta}(s,a) \right] \in \mathbb{R}$.
\end{theorem}

Assuming that the critic error remains small, and with sufficient representation power in the function approximator class, the term $L_{\hat{A}_{\theta}}$ in theorem \ref{thm:MC_MSE} can be made relatively small. We then see that most of the resulting variance can then be annihilated by increasing the number of sampled actions $N_s$ in the all-action Monte Carlo integration estimator.

In order to better understand the effect of using an approximate critic in AAPG, we provide a similar bound in theorem \ref{thm:REINFORCE_MSE} on the MSE for the classical LR (REINFORCE) based on rollouts only.
\begin{theorem}
\label{thm:REINFORCE_MSE}
Let the MSE of the REINFORCE estimator to the true policy gradient be:
\begin{equation*}
L_{R} = MSE_{\pi_{\theta}} \left[ \nabla_{\theta} \log \pi_{\theta}(a|s)  \left( \sum_{k=0}^{T} \gamma^k r(s_k,a_k,s_{k+1}) - \hat{V}_{\theta}(s_k) \right)\right] \enspace .
\end{equation*}
In addition to the regularity assumption of Theorem \ref{thm:MC_MSE}, assume that
$
    \mathbb{E}_{\pi_{\theta}} \left[ \left(G_t - \hat{V}_{\theta}(s_0) - A_{\theta}(s_0,a_0) \right)^2 \right] \leq \xi \in \mathbb{R}
$.
Then $L_R$ satisfies
$
    L_R \leq L + M \xi
$.
\end{theorem}

\paragraph{Remarks} The constant $\xi$ in theorem \ref{thm:REINFORCE_MSE} can be understood as the error in the observed rewards, which is a function of the randomness of the reward distribution and that of the policy. Overall, the variance of the REINFORCE estimator is a function of 1) The randomness due to sampling states and actions, even when $Q_{\theta}$ is perfectly known ($L$); 2) The sensitivity of the policy to variations of $\theta$ ($M$); 3) The stochasticity of the rewards that the policy samples from the environment ($\xi$). As $N_S$ increases, the estimator $\hat{Z}_{N_S}^{\hat{A}_{\theta}}$ becomes more efficient at reducing the mean squared error of the policy gradient compared to the \textit{single-action} REINFORCE estimator $\hat{Z}_R$. This property is observed empirically in section \ref{ss:grad_MSE} where the error is shown to decrease almost perfectly as $1/N_s$.

\section{Convergence of Policy Gradient Methods with a Biased Critic}

Policy gradients belong to the class of stochastic gradient methods for which convergence results are often  \citep{bertsekas2016nonlinear} stated only in the unbiased case. Hence, for likelihood ratio methods of the style of REINFORCE, classical convergence results apply readily \citep{Glynn1987,Lecuyer1991}. In order to address the function approximation case, \citep{Sutton1999} show that if the critic is linear in the so-called \textit{compatible features} then the resulting policy gradient estimator is unbiased. However, the compatible features condition is rather stringent and cannot be satisfied easily beyond the linear case. Because we consider all-action estimators based on approximate critic which can be linear or nonlinear, we need a more general result. Theorem \ref{thm:biased_SGD} characterizes the expected dynamics of the stochastic ascent procedure arising from policy gradient estimators using an approximate critic.

\begin{theorem}
\label{thm:biased_SGD}
    Assume that $\theta \mapsto J(\theta)$ is a $\mu$-Lipschitz smooth concave function ($||\nabla J(x) -\nabla J(y)||_{2}\leq \mu||x-y||_{2}$ for all $x,y$). If the step size at step $k$, $\delta_k$, satisfies $\delta_k \leq \frac{1}{\mu}$, then in expectation:
    \begin{alignat*}{1}
        J(\theta_{k+1}) &\geq J(\theta_{k}) + \frac{1}{2}||\nabla J(\theta_{k})||^2 \\
        &+ \frac{1}{\mu} \left[ \left(\nabla J(\theta_k) - \frac{1}{2} \text{Bias}\left( \hat{\nabla} J(\theta_k)\right) \right) \cdot \text{Bias}\left( \hat{\nabla} J(\theta_k)\right) - \text{Tr}\left(\text{Var}\left( \hat{\nabla} J(\theta_k)\right)\right) \right]
    \end{alignat*}
    Where $\hat{\nabla}J(\theta_k)$ is any (potentially biased) estimator of $\nabla J(\theta_k)$ and $\theta_{k+1} = \theta_{k} + \delta_k \hat{\nabla}J(\theta_k)$.
\end{theorem}

\paragraph{Remarks} This result highlights several central aspects of our methods: 1) Reducing the variance of the gradient estimator is beneficial in terms of sample efficiency; 2) Having a biased gradient estimator can either increase or decrease the convergence speed, depending on the sign of the dot product of the bias and the true gradient; 3) Assuming that the bias of the gradient estimator remains small compared to the true gradient, monotonic improvements can still be guaranteed in expectation.

\section{Experimental Results}

In this section, we demonstrate the use of the all-action policy gradient (AAPG) algorithms on a variety of MuJoCo tasks \citep{Todorov2012}. We used a Gaussian parameterization for the policy whose mean is the output of a neural network, and whose covariance matrix is diagonal and fixed (any other distribution could have been used, such as Beta or $\chi^2$). Both the mean and the actions sampled from the policy are clipped to the admissible range of the action space. We learn both $\hat{Q}$ and $\hat{V}$ using neural networks. The $\hat{V}$ function is learned by least-squares regression on the discounted returns while $\hat{Q}$ is learned by expected SARSA \citep{vanSeijen2009}. It is essential to note that our experiment setup makes REINFORCE (with baseline) and our methods fully comparable since the only difference is in how the policy gradients are computed. All hyperparameters are the same for all algorithms on a given task. More implementation details are given in the supplementary material.
\begin{figure}[]
    \centering
    \resizebox{0.325\textwidth}{!}{\input{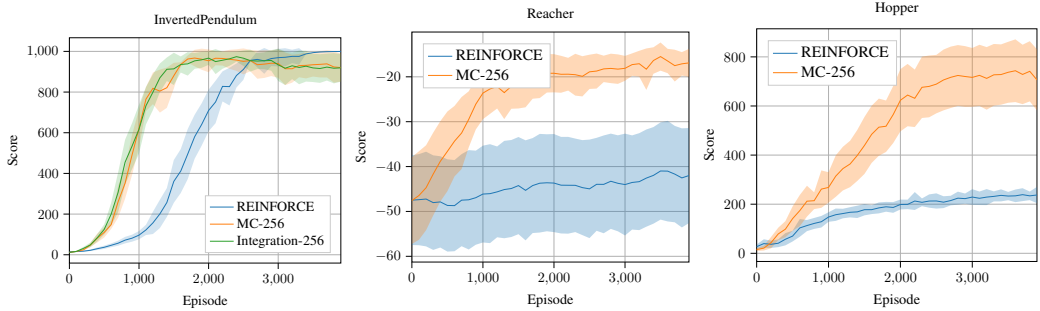}}
    \resizebox{0.325\textwidth}{!}{
\begin{tikzpicture}

\definecolor{color0}{rgb}{0.12156862745098,0.466666666666667,0.705882352941177}
\definecolor{color1}{rgb}{1,0.498039215686275,0.0549019607843137}

\begin{axis}[
legend cell align={left},
legend style={at={(0.5,0.95)},draw=white!80.0!black,nodes={scale=1.1, transform shape}},
tick align=outside,
tick pos=left,
title={Reacher},
x grid style={white!69.01960784313725!black},
xlabel={Episode},
xmajorgrids,
xmin=0, xmax=3900,
xtick style={color=black},
y grid style={white!69.01960784313725!black},
ylabel={Score},
ymajorgrids,
ymin=-61.279034744789, ymax=-10.0178832098541,
ytick style={color=black}
]
\path [fill=color0, fill opacity=0.3]
(axis cs:0,-37.4768939448908)
--(axis cs:0,-57.4871682845727)
--(axis cs:100,-57.531066457563)
--(axis cs:200,-57.7879705559295)
--(axis cs:300,-58.5731134979854)
--(axis cs:400,-58.04753909274)
--(axis cs:500,-58.9489824022919)
--(axis cs:600,-58.8674117498847)
--(axis cs:700,-57.7505807894715)
--(axis cs:800,-58.4763371225593)
--(axis cs:900,-57.4157373745473)
--(axis cs:1000,-56.9169754317362)
--(axis cs:1100,-57.0616630618669)
--(axis cs:1200,-55.7590562729229)
--(axis cs:1300,-55.9761912029487)
--(axis cs:1400,-55.7021589836378)
--(axis cs:1500,-55.0656348491011)
--(axis cs:1600,-56.4489399284141)
--(axis cs:1700,-54.8160982992848)
--(axis cs:1800,-54.6997399553945)
--(axis cs:1900,-54.4221062389907)
--(axis cs:2000,-54.6093639477567)
--(axis cs:2100,-54.9097143427724)
--(axis cs:2200,-54.9538445270378)
--(axis cs:2300,-54.6002848643126)
--(axis cs:2400,-55.2780070277376)
--(axis cs:2500,-55.9461391034771)
--(axis cs:2600,-54.3144155548532)
--(axis cs:2700,-54.8644470062172)
--(axis cs:2800,-53.8897644049779)
--(axis cs:2900,-54.4290964803982)
--(axis cs:3000,-55.3571134059206)
--(axis cs:3100,-54.0385500893149)
--(axis cs:3200,-54.1669143017351)
--(axis cs:3300,-53.403245831977)
--(axis cs:3400,-52.7272539380068)
--(axis cs:3500,-51.8207401137552)
--(axis cs:3600,-52.2268232283664)
--(axis cs:3700,-52.4235674029458)
--(axis cs:3800,-53.5710685114652)
--(axis cs:3900,-52.5990181263558)
--(axis cs:3900,-31.4076046905367)
--(axis cs:3900,-31.4076046905367)
--(axis cs:3800,-31.480233859266)
--(axis cs:3700,-30.9337873527366)
--(axis cs:3600,-29.8027490980949)
--(axis cs:3500,-30.1511422977071)
--(axis cs:3400,-31.1996061889538)
--(axis cs:3300,-31.7347994147253)
--(axis cs:3200,-32.5587078914421)
--(axis cs:3100,-33.0485861237801)
--(axis cs:3000,-32.6618941601664)
--(axis cs:2900,-33.0318114720294)
--(axis cs:2800,-32.6555263630268)
--(axis cs:2700,-33.0654655998628)
--(axis cs:2600,-33.5966048668051)
--(axis cs:2500,-33.9858124057053)
--(axis cs:2400,-34.1056284287301)
--(axis cs:2300,-33.7597700049309)
--(axis cs:2200,-33.3804546792066)
--(axis cs:2100,-33.386526246541)
--(axis cs:2000,-32.7473497059042)
--(axis cs:1900,-32.8031819109636)
--(axis cs:1800,-32.7163828442936)
--(axis cs:1700,-33.6493192961081)
--(axis cs:1600,-34.1480565171269)
--(axis cs:1500,-33.4111694510229)
--(axis cs:1400,-34.0787358885144)
--(axis cs:1300,-34.2737925983256)
--(axis cs:1200,-35.518037437199)
--(axis cs:1100,-35.0121048374794)
--(axis cs:1000,-35.3384592356849)
--(axis cs:900,-36.4523982220869)
--(axis cs:800,-36.3329931928884)
--(axis cs:700,-37.2244084087201)
--(axis cs:600,-38.544009436132)
--(axis cs:500,-38.380643425291)
--(axis cs:400,-37.7008783279194)
--(axis cs:300,-37.4967391498642)
--(axis cs:200,-36.6814589640475)
--(axis cs:100,-37.2211108524062)
--(axis cs:0,-37.4768939448908)
--cycle;

\path [fill=color1, fill opacity=0.3]
(axis cs:0,-37.8883129607016)
--(axis cs:0,-57.2634025088766)
--(axis cs:100,-56.3041239192708)
--(axis cs:200,-54.1825405344794)
--(axis cs:300,-50.9032224585532)
--(axis cs:400,-47.7918755123999)
--(axis cs:500,-45.4908600948311)
--(axis cs:600,-42.1934557161543)
--(axis cs:700,-40.5531211259652)
--(axis cs:800,-35.7402300594294)
--(axis cs:900,-31.732856188598)
--(axis cs:1000,-29.4428305550726)
--(axis cs:1100,-28.2563856556842)
--(axis cs:1200,-27.0632774628657)
--(axis cs:1300,-29.4157428357355)
--(axis cs:1400,-27.1183696598035)
--(axis cs:1500,-27.0567676086558)
--(axis cs:1600,-26.7921967368587)
--(axis cs:1700,-24.881094879418)
--(axis cs:1800,-24.7356662372486)
--(axis cs:1900,-23.5044578594595)
--(axis cs:2000,-23.4600968597126)
--(axis cs:2100,-23.5485528330622)
--(axis cs:2200,-23.0661243329984)
--(axis cs:2300,-24.1888213107644)
--(axis cs:2400,-24.3674699092646)
--(axis cs:2500,-22.705443165282)
--(axis cs:2600,-22.6212262065657)
--(axis cs:2700,-21.1751234281439)
--(axis cs:2800,-21.4056554323926)
--(axis cs:2900,-21.5716043588106)
--(axis cs:3000,-20.8751248607747)
--(axis cs:3100,-19.7821664407681)
--(axis cs:3200,-19.9593733188649)
--(axis cs:3300,-22.9189232814769)
--(axis cs:3400,-19.3173598402227)
--(axis cs:3500,-18.6268292941399)
--(axis cs:3600,-19.1860063445685)
--(axis cs:3700,-21.0516540488979)
--(axis cs:3800,-19.7211350566892)
--(axis cs:3900,-20.0039218809586)
--(axis cs:3900,-13.7854129833788)
--(axis cs:3900,-13.7854129833788)
--(axis cs:3800,-14.4919656505865)
--(axis cs:3700,-13.9626011221582)
--(axis cs:3600,-13.7617545452431)
--(axis cs:3500,-12.3479355523512)
--(axis cs:3400,-13.4168988063779)
--(axis cs:3300,-13.6784184819265)
--(axis cs:3200,-14.1999487182528)
--(axis cs:3100,-14.7727546221677)
--(axis cs:3000,-15.3319859356272)
--(axis cs:2900,-15.0008274098037)
--(axis cs:2800,-14.8455788293171)
--(axis cs:2700,-15.2853327426741)
--(axis cs:2600,-14.8104143307724)
--(axis cs:2500,-15.0692723076192)
--(axis cs:2400,-15.403727787287)
--(axis cs:2300,-14.8679559824833)
--(axis cs:2200,-15.7653216285078)
--(axis cs:2100,-15.3193515749988)
--(axis cs:2000,-14.9520752260392)
--(axis cs:1900,-14.9222420498652)
--(axis cs:1800,-15.1022364494075)
--(axis cs:1700,-15.630864512157)
--(axis cs:1600,-15.7152175755419)
--(axis cs:1500,-14.8695687108429)
--(axis cs:1400,-17.12412943942)
--(axis cs:1300,-17.6173963134169)
--(axis cs:1200,-17.3000363224397)
--(axis cs:1100,-17.1368603345611)
--(axis cs:1000,-17.8070157470968)
--(axis cs:900,-20.433521122661)
--(axis cs:800,-23.2355674471987)
--(axis cs:700,-24.4943758045492)
--(axis cs:600,-26.4838218225083)
--(axis cs:500,-27.7266161388062)
--(axis cs:400,-29.8920542719715)
--(axis cs:300,-32.4428614256408)
--(axis cs:200,-35.2488051318144)
--(axis cs:100,-36.5838055189604)
--(axis cs:0,-37.8883129607016)
--cycle;

\addplot [semithick, color0]
table {%
0 -47.4820311147318
100 -47.3760886549846
200 -47.2347147599885
300 -48.0349263239248
400 -47.8742087103297
500 -48.6648129137915
600 -48.7057105930084
700 -47.4874945990958
800 -47.4046651577239
900 -46.9340677983171
1000 -46.1277173337105
1100 -46.0368839496731
1200 -45.6385468550609
1300 -45.1249919006372
1400 -44.8904474360761
1500 -44.238402150062
1600 -45.2984982227705
1700 -44.2327087976964
1800 -43.708061399844
1900 -43.6126440749772
2000 -43.6783568268305
2100 -44.1481202946567
2200 -44.1671496031222
2300 -44.1800274346218
2400 -44.6918177282339
2500 -44.9659757545912
2600 -43.9555102108291
2700 -43.96495630304
2800 -43.2726453840024
2900 -43.7304539762138
3000 -44.0095037830435
3100 -43.5435681065475
3200 -43.3628110965886
3300 -42.5690226233511
3400 -41.9634300634803
3500 -40.9859412057311
3600 -41.0147861632307
3700 -41.6786773778412
3800 -42.5256511853656
3900 -42.0033114084462
};
\addlegendentry{REINFORCE}
\addplot [semithick, color1]
table {%
0 -47.5758577347891
100 -46.4439647191156
200 -44.7156728331469
300 -41.673041942097
400 -38.8419648921857
500 -36.6087381168186
600 -34.3386387693313
700 -32.5237484652572
800 -29.4878987533141
900 -26.0831886556295
1000 -23.6249231510847
1100 -22.6966229951226
1200 -22.1816568926527
1300 -23.5165695745762
1400 -22.1212495496118
1500 -20.9631681597493
1600 -21.2537071562003
1700 -20.2559796957875
1800 -19.9189513433281
1900 -19.2133499546623
2000 -19.2060860428759
2100 -19.4339522040305
2200 -19.4157229807531
2300 -19.5283886466239
2400 -19.8855988482758
2500 -18.8873577364506
2600 -18.715820268669
2700 -18.230228085409
2800 -18.1256171308549
2900 -18.2862158843071
3000 -18.1035553982009
3100 -17.2774605314679
3200 -17.0796610185588
3300 -18.2986708817017
3400 -16.3671293233003
3500 -15.4873824232456
3600 -16.4738804449058
3700 -17.5071275855281
3800 -17.1065503536379
3900 -16.8946674321687
};
\addlegendentry{MC-256}
\end{axis}

\end{tikzpicture}}
    \resizebox{0.325\textwidth}{!}{
\begin{tikzpicture}

\definecolor{color0}{rgb}{0.12156862745098,0.466666666666667,0.705882352941177}
\definecolor{color1}{rgb}{1,0.498039215686275,0.0549019607843137}

\begin{axis}[
legend cell align={left},
legend style={at={(0.5,0.95)},draw=white!80.0!black,nodes={scale=1.1, transform shape}},
tick align=outside,
tick pos=left,
title={Hopper},
x grid style={white!69.01960784313725!black},
xlabel={Episode},
xmajorgrids,
xmin=0, xmax=3900,
xtick style={color=black},
y grid style={white!69.01960784313725!black},
ylabel={Score},
ymajorgrids,
ymin=-33.2539627254558, ymax=914.390329297215,
ytick style={color=black}
]
\path [fill=color0, fill opacity=0.3]
(axis cs:0,35.6194537552699)
--(axis cs:0,18.3212987792029)
--(axis cs:100,21.9390196276717)
--(axis cs:200,21.9212281677561)
--(axis cs:300,25.0496608559943)
--(axis cs:400,32.274176032777)
--(axis cs:500,48.1285227700883)
--(axis cs:600,67.2578312850565)
--(axis cs:700,90.3099480883879)
--(axis cs:800,99.3436379908903)
--(axis cs:900,104.968191850794)
--(axis cs:1000,128.301811787199)
--(axis cs:1100,131.973394189463)
--(axis cs:1200,142.524073344348)
--(axis cs:1300,146.280756861052)
--(axis cs:1400,143.601450128967)
--(axis cs:1500,154.508983441804)
--(axis cs:1600,159.76082678464)
--(axis cs:1700,162.569332378923)
--(axis cs:1800,169.438247502389)
--(axis cs:1900,164.896923621872)
--(axis cs:2000,178.573158927008)
--(axis cs:2100,179.868199884225)
--(axis cs:2200,176.44882485582)
--(axis cs:2300,191.821480505154)
--(axis cs:2400,185.754927701331)
--(axis cs:2500,182.129266142473)
--(axis cs:2600,184.0637620285)
--(axis cs:2700,195.258167851699)
--(axis cs:2800,198.079133377911)
--(axis cs:2900,201.802654014755)
--(axis cs:3000,197.022141303099)
--(axis cs:3100,196.552212767106)
--(axis cs:3200,196.224767143212)
--(axis cs:3300,211.332795754839)
--(axis cs:3400,211.407674387457)
--(axis cs:3500,216.194036595603)
--(axis cs:3600,208.34137754573)
--(axis cs:3700,212.523386527315)
--(axis cs:3800,216.525368466225)
--(axis cs:3900,204.353067195707)
--(axis cs:3900,270.6248833908)
--(axis cs:3900,270.6248833908)
--(axis cs:3800,251.513165598382)
--(axis cs:3700,265.176240915582)
--(axis cs:3600,259.719853091342)
--(axis cs:3500,250.127048748546)
--(axis cs:3400,260.466195663913)
--(axis cs:3300,252.273300840988)
--(axis cs:3200,261.594621588829)
--(axis cs:3100,251.615417414587)
--(axis cs:3000,261.229211706172)
--(axis cs:2900,243.450919126978)
--(axis cs:2800,251.140453686474)
--(axis cs:2700,234.363981783287)
--(axis cs:2600,232.374830684997)
--(axis cs:2500,244.207057929376)
--(axis cs:2400,239.930692649003)
--(axis cs:2300,223.49455344614)
--(axis cs:2200,249.405395976333)
--(axis cs:2100,218.456652845617)
--(axis cs:2000,218.903918057305)
--(axis cs:1900,208.623927793775)
--(axis cs:1800,209.400482258909)
--(axis cs:1700,207.241456334972)
--(axis cs:1600,195.724537659012)
--(axis cs:1500,200.740961447822)
--(axis cs:1400,193.907265541234)
--(axis cs:1300,187.821904986686)
--(axis cs:1200,181.376280257486)
--(axis cs:1100,182.345150362644)
--(axis cs:1000,167.161248228075)
--(axis cs:900,152.153460192943)
--(axis cs:800,144.018353883753)
--(axis cs:700,135.662622406267)
--(axis cs:600,139.443928721985)
--(axis cs:500,91.9780712128362)
--(axis cs:400,82.1877942818362)
--(axis cs:300,56.7428929821298)
--(axis cs:200,51.5221136769575)
--(axis cs:100,57.1532959827754)
--(axis cs:0,35.6194537552699)
--cycle;

\path [fill=color1, fill opacity=0.3]
(axis cs:0,20.9119886843001)
--(axis cs:0,9.82077782102926)
--(axis cs:100,11.2803392356039)
--(axis cs:200,20.2999313934997)
--(axis cs:300,53.2190205553025)
--(axis cs:400,65.5257038112994)
--(axis cs:500,91.5089389931114)
--(axis cs:600,110.09995579112)
--(axis cs:700,149.38629429537)
--(axis cs:800,179.95276950831)
--(axis cs:900,185.13537330359)
--(axis cs:1000,206.403292284054)
--(axis cs:1100,229.805381008419)
--(axis cs:1200,255.957873127392)
--(axis cs:1300,269.314463821273)
--(axis cs:1400,294.663252684682)
--(axis cs:1500,323.135573918916)
--(axis cs:1600,368.09928282071)
--(axis cs:1700,397.120512697064)
--(axis cs:1800,412.710430895863)
--(axis cs:1900,448.973109640763)
--(axis cs:2000,496.606534477625)
--(axis cs:2100,517.484685473756)
--(axis cs:2200,514.108379519101)
--(axis cs:2300,551.15255528123)
--(axis cs:2400,553.901054887739)
--(axis cs:2500,569.994879114298)
--(axis cs:2600,582.211400315005)
--(axis cs:2700,591.706386839368)
--(axis cs:2800,600.489491757842)
--(axis cs:2900,596.310745237433)
--(axis cs:3000,596.681109754982)
--(axis cs:3100,600.018575301238)
--(axis cs:3200,584.933859273015)
--(axis cs:3300,601.189506749245)
--(axis cs:3400,605.181026240134)
--(axis cs:3500,613.118038854132)
--(axis cs:3600,615.835281544411)
--(axis cs:3700,607.409982333182)
--(axis cs:3800,618.575860414522)
--(axis cs:3900,581.735891024927)
--(axis cs:3900,830.859602064663)
--(axis cs:3900,830.859602064663)
--(axis cs:3800,863.861696632776)
--(axis cs:3700,849.506417964281)
--(axis cs:3600,871.31558875073)
--(axis cs:3500,861.429290730363)
--(axis cs:3400,850.834505084155)
--(axis cs:3300,852.622532668323)
--(axis cs:3200,839.873982763317)
--(axis cs:3100,849.337300603287)
--(axis cs:3000,836.197983546415)
--(axis cs:2900,843.798418845849)
--(axis cs:2800,848.095861979374)
--(axis cs:2700,841.819922643253)
--(axis cs:2600,829.7768692126)
--(axis cs:2500,808.047404590092)
--(axis cs:2400,805.605805759555)
--(axis cs:2300,800.931940904933)
--(axis cs:2200,748.390084433139)
--(axis cs:2100,771.405956049279)
--(axis cs:2000,750.083332116229)
--(axis cs:1900,681.41578097049)
--(axis cs:1800,622.330246217027)
--(axis cs:1700,630.041945667756)
--(axis cs:1600,601.070326746019)
--(axis cs:1500,555.368618908637)
--(axis cs:1400,502.386695606833)
--(axis cs:1300,458.273514501872)
--(axis cs:1200,433.901529134024)
--(axis cs:1100,398.543489041747)
--(axis cs:1000,330.254473667379)
--(axis cs:900,337.666430889282)
--(axis cs:800,248.520287661941)
--(axis cs:700,275.204351763274)
--(axis cs:600,237.381107197062)
--(axis cs:500,190.918578176866)
--(axis cs:400,131.101662027529)
--(axis cs:300,104.207002687657)
--(axis cs:200,66.4518260738776)
--(axis cs:100,30.7131600076301)
--(axis cs:0,20.9119886843001)
--cycle;

\addplot [semithick, color0]
table {%
0 26.9703762672364
100 39.5461578052236
200 36.7216709223568
300 40.8962769190621
400 57.2309851573066
500 70.0532969914623
600 103.350880003521
700 112.986285247328
800 121.680995937322
900 128.560826021869
1000 147.731530007637
1100 157.159272276054
1200 161.950176800917
1300 167.051330923869
1400 168.754357835101
1500 177.624972444813
1600 177.742682221826
1700 184.905394356947
1800 189.419364880649
1900 186.760425707823
2000 198.738538492156
2100 199.162426364921
2200 212.927110416076
2300 207.658016975647
2400 212.842810175167
2500 213.168162035925
2600 208.219296356749
2700 214.811074817493
2800 224.609793532193
2900 222.626786570867
3000 229.125676504636
3100 224.083815090846
3200 228.90969436602
3300 231.803048297914
3400 235.936935025685
3500 233.160542672074
3600 234.030615318536
3700 238.849813721448
3800 234.019267032304
3900 237.488975293254
};
\addlegendentry{REINFORCE}
\addplot [semithick, color1]
table {%
0 15.3663832526647
100 20.996749621617
200 43.3758787336886
300 78.71301162148
400 98.3136829194144
500 141.213758584989
600 173.740531494091
700 212.295323029322
800 214.236528585126
900 261.400902096436
1000 268.328882975717
1100 314.174435025083
1200 344.929701130708
1300 363.793989161573
1400 398.524974145757
1500 439.252096413777
1600 484.584804783364
1700 513.58122918241
1800 517.520338556445
1900 565.194445305626
2000 623.344933296927
2100 644.445320761518
2200 631.24923197612
2300 676.042248093081
2400 679.753430323647
2500 689.021141852195
2600 705.994134763802
2700 716.763154741311
2800 724.292676868608
2900 720.054582041641
3000 716.439546650698
3100 724.677937952262
3200 712.403921018166
3300 726.906019708784
3400 728.007765662144
3500 737.273664792248
3600 743.57543514757
3700 728.458200148732
3800 741.218778523649
3900 706.297746544795
};
\addlegendentry{MC-256}
\end{axis}

\end{tikzpicture}}
    \caption{Learning curves for InvertedPendulum-v1 (left), Reacher-v1 (center), Hopper-v1 (right)}
    \label{fig:learning_curves}
\end{figure}
\subsection{Performance of All-Action Monte Carlo Integration Estimator}

In Figure \ref{fig:learning_curves}, we give learning curves for $3$ MuJoCo continuous control tasks with action spaces of diverse dimensions, averaged across $25$ random initializations. Error bars use Student $90\%$ confidence intervals for the empirical mean at every episode. As an illustration of numerical quadrature methods, we provide runs with trapezoidal integration AAPG for InvertedPendulum-v1, but chose to focus on Monte-Carlo methods in the other experiments since quadrature methods scale poorly with dimension and did not yield interesting improvements over Monte-Carlo even in a simple setting.

\subsection{Gradient MSE and sample efficiency}\label{ss:grad_MSE}

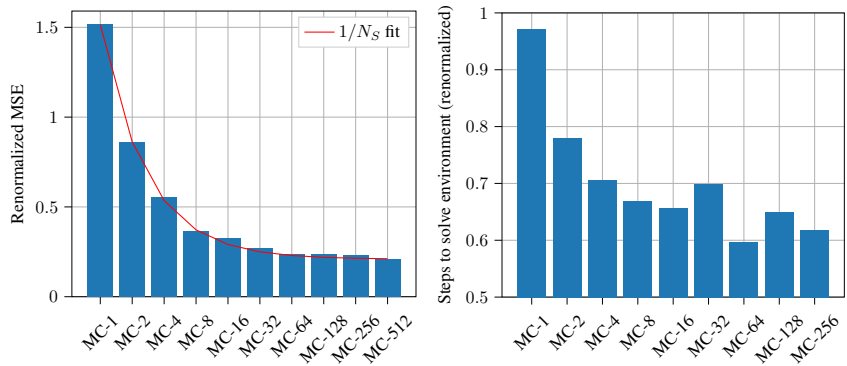
\begin{figure}[]
    \centering
    \resizebox{0.4\textwidth}{!}{
\begin{tikzpicture}

\definecolor{color0}{rgb}{0.12156862745098,0.466666666666667,0.705882352941177}

\begin{axis}[
legend cell align={left},
legend style={draw=white!80.0!black},
tick align=outside,
tick pos=left,
x grid style={white!69.01960784313725!black},
xmajorgrids,
xmin=-0.89, xmax=9.89,
xtick style={color=black},
xtick={0,1,2,3,4,5,6,7,8,9},
xticklabel style = {rotate=45.0},
xticklabels={MC-1,MC-2,MC-4,MC-8,MC-16,MC-32,MC-64,MC-128,MC-256,MC-512},
y grid style={white!69.01960784313725!black},
ylabel={Renormalized MSE},
ymajorgrids,
ymin=0, ymax=1.59067164179104,
ytick style={color=black}
]
\draw[fill=color0,draw opacity=0] (axis cs:-0.4,0) rectangle (axis cs:0.4,1.51492537313433);
\draw[fill=color0,draw opacity=0] (axis cs:0.6,0) rectangle (axis cs:1.4,0.858208955223881);
\draw[fill=color0,draw opacity=0] (axis cs:1.6,0) rectangle (axis cs:2.4,0.552238805970149);
\draw[fill=color0,draw opacity=0] (axis cs:2.6,0) rectangle (axis cs:3.4,0.365671641791045);
\draw[fill=color0,draw opacity=0] (axis cs:3.6,0) rectangle (axis cs:4.4,0.328358208955224);
\draw[fill=color0,draw opacity=0] (axis cs:4.6,0) rectangle (axis cs:5.4,0.26865671641791);
\draw[fill=color0,draw opacity=0] (axis cs:5.6,0) rectangle (axis cs:6.4,0.238805970149254);
\draw[fill=color0,draw opacity=0] (axis cs:6.6,0) rectangle (axis cs:7.4,0.238805970149254);
\draw[fill=color0,draw opacity=0] (axis cs:7.6,0) rectangle (axis cs:8.4,0.23134328358209);
\draw[fill=color0,draw opacity=0] (axis cs:8.6,0) rectangle (axis cs:9.4,0.208955223880597);
\addplot [semithick, red]
table {%
0 1.51492537313433
1 0.861940298507463
2 0.53544776119403
3 0.372201492537313
4 0.290578358208955
5 0.249766791044776
6 0.229361007462687
7 0.219158115671642
8 0.214056669776119
9 0.211505946828358
};
\addlegendentry{$1/{N_S}$ fit}
\end{axis}

\end{tikzpicture}}
    \resizebox{0.4\textwidth}{!}{
\begin{tikzpicture}

\definecolor{color0}{rgb}{0.12156862745098,0.466666666666667,0.705882352941177}

\begin{axis}[
tick align=outside,
tick pos=left,
x grid style={white!69.01960784313725!black},
xmajorgrids,
xmin=-0.84, xmax=8.84,
xtick style={color=black},
xtick={0,1,2,3,4,5,6,7,8},
xticklabel style = {rotate=45.0},
xticklabels={MC-1,MC-2,MC-4,MC-8,MC-16,MC-32,MC-64,MC-128,MC-256},
y grid style={white!69.01960784313725!black},
ylabel={Steps to solve environment (renormalized)},
ymajorgrids,
ymin=0.5, ymax=1,
ytick style={color=black}
]
\draw[fill=color0,draw opacity=0] (axis cs:-0.4,0) rectangle (axis cs:0.4,0.971948372133327);
\draw[fill=color0,draw opacity=0] (axis cs:0.6,0) rectangle (axis cs:1.4,0.780075435656972);
\draw[fill=color0,draw opacity=0] (axis cs:1.6,0) rectangle (axis cs:2.4,0.70551597430781);
\draw[fill=color0,draw opacity=0] (axis cs:2.6,0) rectangle (axis cs:3.4,0.668253490638755);
\draw[fill=color0,draw opacity=0] (axis cs:3.6,0) rectangle (axis cs:4.4,0.656919522488305);
\draw[fill=color0,draw opacity=0] (axis cs:4.6,0) rectangle (axis cs:5.4,0.698501993133185);
\draw[fill=color0,draw opacity=0] (axis cs:5.6,0) rectangle (axis cs:6.4,0.596265205712017);
\draw[fill=color0,draw opacity=0] (axis cs:6.6,0) rectangle (axis cs:7.4,0.648982199896733);
\draw[fill=color0,draw opacity=0] (axis cs:7.6,0) rectangle (axis cs:8.4,0.617028882568677);
\end{axis}

\end{tikzpicture}}
    \caption{Empirical MSE of $\hat{Z}^{\hat{A}_{\theta}}_{N_S}$ for $0\leq \log_2(N_S) \leq 9$ in a given training step on InvertedPendulum-v1 (left) and average number of steps (across $25$ random initializations) to solve the environment. The rightmost figure shows that reducing the MSE of the gradient estimator increases sample efficiency.}
    \label{fig:grad_mse}
\end{figure}

As an empirical verification of Theorem \ref{thm:MC_MSE}, we can use the following approach to compute an approximation of the MSE of the $\hat{Z}^{\hat{A}_{\theta}}_{N_S}$ estimators for various values of $N_S$:
We first train the policy, critic, and baseline for $1000$ episodes on a given environment. Since the REINFORCE estimator is an unbiased estimate of the policy gradient, we use an additional $1000$ rollouts to compute an accurate estimate of the true gradient. We then use this proxy gradient to compute the MSE of the $\hat{Z}^{\hat{A}_{\theta}}_{N_S}$ estimator for various values of $N_S$ by computing the empirical MSE across $1,000$ gradient estimates for each value of $N_S$.
Our results on \textit{InvertedPendulum-v1} are shown in Figure \ref{fig:grad_mse}, and highlight the benefits of sampling more actions for each state when computing the policy gradient estimate. As $N_S$ increases, the MSE converges (with a clearly visible $1/N_S$ progression) to a positive value, which corresponds to the MSE of the learned advantage estimate (first term of Eq. \ref{eq:MSE_bound}). Figure \ref{fig:grad_mse} also draws a parallel between the MSE reduction and the sample efficiency of the algorithm (here defined as the number of steps needed to reach an average score of $950$ over a window of $100$ consecutive episodes).

\section{Conclusion}

We have shown, both in theory and practice, that the all-action policy gradient estimator of \citep{sutton2001} can yield performance gains and better sample efficiency by acting as a variance reduction technique. We have established that the mechanism at play in this approach is that of \textit{conditioning}. We also provided practical algorithms for implementing all-actions estimators in large or continuous action spaces which are compatible with linear or nonlinear function approximation.  We derived novel bounds on the MSE of our Monte-Carlo integration estimator and compared it both theoretically and experimentally to that of the REINFORCE estimator. Our theoretical analysis also provides the first characterization of the dynamics of policy gradient methods with a biased critic.

Our methods could be extended in many ways, notably by combining them with other variance reduction techniques such as control variates. As pointed out by \citep{Glynn2002}, the conditional expectation term used in the conditioning technique can itself be used as a control variate. In other words, efforts spent computing the integrand by numerical integration can then be re-used to form a complementary control variate. Note finally that given the basis of our Monte Carlo integration all-action estimator in importance sampling, it is natural to think of this approach as a form of off-policy policy gradient method \citep{Degris2012} where the \textit{off-policiness} is only inside the inner expectation \eqref{eq:originalPG2} rather than the outer expectation over the stationary distribution \citep{Yao2019}. Hence, additional variance reductions could be obtained by choosing a better sampling distribution beyond $\pi_\theta$ itself \citep{Lecuyer1994}.

\newpage

\bibliography{references}
\bibliographystyle{unsrtnat}

\end{document}